\documentclass[10pt,twocolumn,letterpaper]{article}
\usepackage[top=0.75in, bottom=1in, left=0.75in, right=0.75in]{geometry}
\usepackage{times}
\usepackage{authblk}

\usepackage{microtype}
\usepackage{graphicx}
\usepackage{booktabs}

\usepackage{enumitem}
  \newlist{inlinelist}{enumerate*}{1}
  \setlist*[inlinelist,1]{%
          label=(\roman*),
      }

\usepackage{adjustbox}
\usepackage{numprint}
\usepackage{multirow}

\usepackage{tikz}
\usetikzlibrary{tikzmark}
\newcommand{\tikzxmark}{%
\tikz[scale=0.23] {
    \draw[line width=0.7,line cap=round] (0,0) to [bend left=6] (1,1);
    \draw[line width=0.7,line cap=round] (0.2,0.95) to [bend right=3] (0.8,0.05);
}}
\newcommand{\tikzcmark}{%
\tikz[scale=0.23] {
    \draw[line width=0.7,line cap=round] (0.25,0) to [bend left=10] (1,1);
    \draw[line width=0.8,line cap=round] (0,0.35) to [bend right=1] (0.23,0);
}}

\usepackage{amsmath}
\usepackage{amssymb}
\usepackage{mathtools}
\usepackage{amsthm}
\usepackage{nicefrac}

\newtheorem{lemma}{Lemma}

\usepackage[numbers]{natbib}
\usepackage[hyphens]{url}
\usepackage{hyperref}
\usepackage{breakurl}
\usepackage{xurl}

\usepackage[capitalize,nameinlink]{cleveref}
\usepackage{colortbl}
\usepackage{xcolor}
\usepackage{algorithmic}
\usepackage{array}
\usepackage{authblk}

\usepackage{url}
\begin{document}
\title{Efficient and Scalable Implementation of Differentially Private Deep Learning without Shortcuts \thanks{This work
has been accepted for publication at the IEEE Conference on Secure and
Trustworthy Machine Learning (SaTML'26). The final version will be
available on IEEE Xplore.}}

\author[1]{Sebastian Rodriguez Beltran}
\author[1,3]{Marlon Tobaben}
\author[1]{Joonas Jälkö}
\author[2]{Niki Loppi}
\author[1]{Antti Honkela}
\affil[1]{Department of Computer Science,
University of Helsinki, Finland}
\affil[2]{NVIDIA}
\affil[3]{CSC – IT Center for Science, Finland}

\date{}
\maketitle

\begin{abstract}
Differentially private stochastic gradient descent (DP-SGD) is the standard algorithm for training machine learning models under differential privacy (DP). The most common DP-SGD privacy accountants rely on Poisson subsampling to ensure the theoretical DP guarantees. Implementing computationally efficient DP-SGD with Poisson subsampling is not trivial, which leads many implementations to taking a shortcut by using computationally faster subsampling. We quantify the computational cost of training deep learning models under DP by implementing and benchmarking efficient methods with the correct Poisson subsampling. We find that using the naive implementation of DP-SGD with Opacus in PyTorch has a throughput between 2.6 and 8 times lower than that of SGD. However, efficient gradient clipping implementations like Ghost Clipping can roughly halve this cost. We propose an alternative computationally efficient implementation of DP-SGD with JAX that uses Poisson subsampling and performs comparably with efficient clipping optimizations based on PyTorch. We study the scaling behavior using up to 80 GPUs and find that DP-SGD scales better than SGD.\looseness-1
\end{abstract}

\section{Introduction}
Differential Privacy~(DP)~\citep{dwork2006calibrating} is the gold standard for formalizing the privacy leakage of training data in Machine Learning (ML) and mitigating the risk of privacy attacks~\citep{balle2022reconstructing,carlini2021extracting} on the training data. 
The established algorithm for integrating DP into the training pipeline of deep learning models is DP stochastic gradient descent (DP-SGD)~\citep{dp-sgd-rajkumar-2012,dp-sgd-song-2013,dp-sgd-abadi-2016}, which is the DP adaptation of SGD. 
In order for the theoretical DP guarantees to hold, implementations of DP-SGD need to exactly match the theory. 
A particular sampling method called Poisson subsampling, where each example is selected independently at each iteration with a fixed probability, has been extensively studied under DP.
Using Poisson subsampling, the resulting minibatches can be of different sizes, making efficient implementation more difficult. As a result, many existing implementations forego proper implementation of Poisson subsampling, which means that the stated privacy guarantees do not apply.\looseness-1

\begin{figure}[htb]
\begin{center}
    \includegraphics[width=\linewidth]{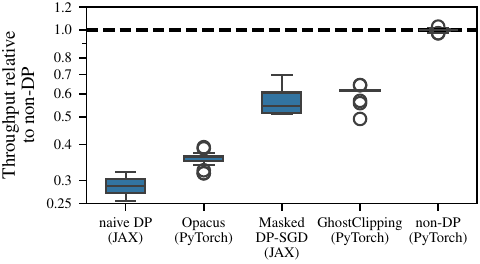}
    \caption{Relative throughput (FP32) to the respective non private baseline (higher is better) on NVIDIA A100. For each optimization method and each model size, we divide its throughput with the non-private counterpart. Throughput is the number of processed instances per second. In this benchmark we distinguish between precision modes. They are available on both frameworks and significantly improve the throughput for the different DP-SGD implementations.}
    \label{fig:MainFigure}
\end{center}
\end{figure}

We focus on the computational efficiency of DP-SGD when implementing it with Poisson subsampling. We re-implement efficient DP-SGD methods with correct Poisson subsampling and benchmark them (see \cref{fig:MainFigure}). We observe that naively implementing Poisson subsampling with JAX leads to re-compilations and we propose a method that avoids this problem, enabling us to utilize JAX compilation advantages for an efficient DP-SGD training. We further study other ways of speeding up the process such as faster clipping and lower precision.

Especially papers using JAX~\citep{de2022unlocking,berrada2023fairness,kurakin2022towards} or ones that build on older versions of the PyTorch library opacus~\citep{li2022large,FastDPbu2022differentially,PrivateVision} take a shortcut and shuffle the data instead of properly implementing Poisson subsampling.
No valid theoretical privacy bounds have been published for these subsampling methods.
Recent research \citep{lebeda2025avoiding,chua2024private,chua2024scalable,annamalai2024shuffleshuffleauditingdpsgd} has shown that they are likely to enjoy far weaker privacy than the Poisson subsampling analysis indicates, leading to privacy risks when using such implementations.

\textbf{List of contributions} In this work we conduct an extensive empirical study on the computational efficiency of DP-SGD using Poisson sub-sampling, focusing on fine-tuning a wide range of large image classification models. Our findings can be applied to any other large models trained or fine-tuned with DP-SGD. Our main contributions are as follows:

\begin{enumerate}[leftmargin=*,noitemsep,topsep=0pt]
    \item We re-implement all DP-SGD methods with Poisson subsampling that is fully DP and share the source code\footnote{\url{https://github.com/DPBayes/Towards-Efficient-Scalable-Training-DP-DL}}.
    \item We propose a JAX implementation relying on proper Poisson sampling that is in comparison to a naive JAX implementation not prone to re-compilation and outperforms its throughput by two times (See \cref{sec:masked-dp-sgd}).
    \item  We find that non-optimized training with DP-SGD costs per-epoch between 2.6 and 3.2 times more than non-private training for ViT and 4 to 8 times for ResNets (See \cref{sec:dpcost}).
    We identify the reasons that lead to the higher computational cost of DP-SGD using profiling.
    \item We benchmark different strategies that can drastically reduce this cost:
    \begin{inlinelist}
        \item More efficient gradient clipping implementations of DP-SGD (See \cref{fig:MainFigure,sec:exp_optimized_dp-sgd}).
        \item Lower Precision with TF32 (See \cref{subsec:lower}).
    \end{inlinelist}

    \item We scale up the training to 80 GPUs and find that DP-SGD scales better than non-private training (See \cref{sec:multgpu}).
\end{enumerate}

\section{Background}

This section will explain the main DP-SGD algorithm and optimizations to alleviate its computational cost.

\subsection{DP-SGD Algorithm}
\cref{alg:dpsgd} is the original DP-SGD algorithm, with virtual batching, as proposed by \citet{dp-sgd-abadi-2016}. DP-SGD has two major drawbacks in comparison to SGD: higher computational cost and loss in utility. DP-SGD requires more memory and is computationally more expensive due to the per-example clipping. The utility in comparison to non-DP training drops, but this can be mitigated to some extent by using larger batch sizes, to mitigate the noise addition ~\citep{räisä2024subsampling}, and training longer~\citep{HowToDP-fy} which further increase the computational cost.

\begin{figure}[h]
\begin{algorithmic}[1]
   \STATE {\bfseries Input:} Training data points $D = \left\{x_{1},\dots,x_{N}\right\}$,\\ loss function $\mathcal{L}(\theta) = \frac{1}{N}\sum_i \mathcal{L}(\theta,x_i)$
   \STATE {\bfseries Parameters:} noise scale $\sigma$, gradient norm bound $C$, number of steps $T$, expected logical batch size $L$,\\physical batch size $p$.
    \STATE {\bfseries Start} 
   
   \FOR{$t \in [T]$}

   \STATE // {Sample batch with Poisson sampling with rate $\nicefrac{L}{N}$}
   \STATE $B \gets \{x_i,\ldots\}$
   \STATE // {Divide $B$ into physical batches of size $p$}.
   \STATE $P \gets \{B_1,\ldots,B_k\}$
   \STATE $\theta_{acc} \gets \mathbf{0}$ 
   \FOR{$s \in P$}
   \FOR{$x_i \in s$}
   \STATE // {Compute gradient}
   \STATE $\textbf{g}_{t}(x_{i}) \leftarrow \nabla_{\theta_{t}} \mathcal{L}(\theta_{t},x_{i})$ 
   \STATE // {Clip gradient}
   \STATE $\overline{\mbox{\textbf{g}}}_t (x_{i}) \leftarrow \textbf{g}_t / \text{max} (1,\frac{\|\textbf{g}_t (x_i) \|_2}{C})  $ 
   \ENDFOR
   \STATE // {Accumulate gradient}
   \STATE $\theta_{acc}  \leftarrow \theta_{acc}+ \sum_{i} \overline{\mbox{\textbf{g}}}_t (x_{i})$ 
   \ENDFOR
   \STATE // {Add noise}
   \STATE $\widetilde{\textbf{g}}_{t} \leftarrow \frac{1}{L}(\theta_{acc} + \mathcal{N}(0,\sigma^{2}C^{2}\textbf{I}))$  
   \STATE // {Take step}
   \STATE $\theta_{t+1} \leftarrow \text{OptimizerUpdate(}\theta_{t}, \widetilde{\textbf{g}}_{t}$)
   \ENDFOR
   \STATE{\bfseries Return} Learned parameters $\theta_T$ and the privacy cost from a privacy accountant.
\end{algorithmic}
   \caption{A variant to the DP-SGD algorithm~\citep{dp-sgd-rajkumar-2012,dp-sgd-song-2013,dp-sgd-abadi-2016} that supports virtual batching of logical batches of size $b$ into smaller phyiscal batches of size $p$ that can be processed in memory.}   \label{alg:dpsgd}

\end{figure}

\textbf{Virtual Batching} distinguishes between logical and physical batches. Logical batches are divided into multiple physical batches to enable optimizer steps with many samples without running out of memory. For instance, we typically sample logical batch sizes of $L = 25000$ while the memory fits $< 300$ samples at once. Implementing DP-SGD with virtual batching (\cref{alg:dpsgd}) does not modify the privacy accounting. The amount of noise added is the same and does not affect the model utility~\citep{HowToDP-fy}. \looseness -1

\textbf{Poisson subsampling} Interestingly, \citet{PrivateVision} and \citet{FastDPbu2022differentially} never mention Poisson subsampling in their works of Mix Ghost clipping and Book Keeping. Furthermore, \citet{PrivateVision} claim a speed-up of $\times 1.7$ against other algorithms with a fixed batch size, which would affect the privacy accounting method. The same happens in practice for JAX implementations \citep{de2022unlocking,berrada2023fairness}, where sampling is done by shuffling the dataset and using each sample once per epoch. While this makes efficient implementation easier, it does not use the correct Poisson subsampling assumed by privacy accounting methods. Therefore, the implementation might have significantly weaker privacy properties than claimed~\citep{lebeda2025avoiding,chua2024private,chua2024scalable,annamalai2024shuffleshuffleauditingdpsgd}. All our experiments are based on Poisson subsampling which is compliant with the commonly used privacy accounting. 

\subsection{DP-SGD Gradient Clipping Optimizations \label{subsec:optim}}

We benchmark five types of clipping methods. \cref{tab:algo} shows which clipping optimizations we are benchmarking against the library or framework that implements it.

\begin{table*}[!h]
    \caption{Benchmarked DP-SGD frameworks and libraries. Note that Opacus Ghost Clipping is in development.}
    \label{tab:algo} 
    \begin{center}
        \begin{small}
            \begin{sc}
            {
            \begin{adjustbox}{max width=\textwidth}
            \arrayrulecolor{black!30}
                \begin{tabular}{l|c|c|c|c|c|c}
                    \toprule
                    &
                    \multicolumn{4}{c|}{PyTorch} & \multicolumn{2}{c}{JAX}\\
                    Clipping Mode & native & Opacus & PrivateVision (PV) & FastDP (BK) & native & ours\\ 
                     & & \citep{OpacusPaper} & \citep{PrivateVision} & \citep{FastDPbu2022differentially} &  & \\
                    \midrule
                         Non-private & \tikzcmark & & & & \tikzcmark  \\\hline
                         Per-example & & \tikzcmark & & & \tikzcmark \\\hline
                         Ghost Clipping~\citep{li2022large} & & (\tikzcmark) &  \tikzcmark & \tikzcmark & \\\hline
                         Mix Ghost~\citep{PrivateVision} & & &\tikzcmark & \tikzcmark & \\\hline
                         Mix Opt ~\citep{FastDPbu2022differentially}  & & & & \tikzcmark& \\\hline
                         Masked DP-SGD (ours, \cref{sec:masked-dp-sgd}) & & & & & & \tikzcmark \\
                    \bottomrule
                    \end{tabular}
                    \end{adjustbox}}
            \end{sc}
        \end{small}
    \end{center}  
\end{table*}

\textbf{Ghost clipping} computes the loss gradient norm  after the backpropagation step and then reweights the loss to update the clipped gradients. Its main contribution is memory savings at the cost of adding another backward pass \citep{li2022large}.

\textbf{Mixed Ghost clipping}~\citep{PrivateVision} is a method that builds on-top of Ghost clipping. It implements the ghost clipping technique for convolutional layers. Its main contribution is that the algorithm will decide when to clip the gradients using per-example or ghost. This difference matters because the ghost clipping is less efficient when the layer's input dimensions are too high-dimensional. E.g., for ResNets, each clipping method will be applied for half of the layers. The first layers will be clipped using the the per-example and then ghost clipping in the bottom layers. As the model goes deeper, the feature size decreases, and the number of channels increases, prioritizing ghost clipping \citep{PrivateVision}.\looseness-1

\textbf{Book Keeping}~\citep{FastDPbu2022differentially} uses all the previous techniques but requires only one backpropagation pass without explicitly calculating the per-example gradients. It avoids the second pass that ghost clipping does by reusing the intermediate results of the output gradients to calculate the sum of the clipped gradients and the clipping factor. Book Keeping can also be implemented together with the Mix Optimization, which does the same evaluation as the mix ghost clipping, but also determines whether doing a second backward pass is more efficient.\looseness-1

\subsection{Batch Sampling Implementations}
As discussed in the introduction, the implementations for all components of the DP-SGD algorithm need to match the DP theory so that the DP privacy guarantees apply. 

Privacy accounting commonly relies on so-called amplification through subsampling where at every iteration a subset of the data is used for training. There are different subsampling methods for a dataset $D$ with size $|D| = N$ and subsampling ratio $q = \nicefrac{L}{N}$:

\begin{enumerate}
    \item Poisson subsampling: from batch $B$ by including each $x \in D$ independently w.p. $q$. The expected size of $B$ is $N q$.\looseness-1
    \item Without replacement subsampling: $B \sim \mathrm{WOR}(D, L)$ with all batches $B$ being of constant size $L$.
    \item Shuffling: Permute $D$, split into disjoint batches of fixed size  $L$ (there are different implementations of this, see \citet{annamalai2024shuffleshuffleauditingdpsgd}).
\end{enumerate}

Both the shuffling procedure and the without replacement subsampling result in batches of equal size which is easier to implement efficiently, e.g., in JAX, and have computational benefits~\citep{HowToDP-fy,annamalai2024shuffleshuffleauditingdpsgd}. However, they have different implications on the privacy guarantees.

\citet{lebeda2025avoiding} showed that without replacement subsampling can result in significantly weaker privacy guarantees than Poisson subsampling when you perturb the gradients with a fixed noise std. They observed for some realistically chosen hyperparameters that without replacement subsampling results in $\epsilon>10$ while the Poisson subsampling would result in $\epsilon=1$. 

\citet{annamalai2024shuffleshuffleauditingdpsgd} focus on shuffling and audit some state-of-the-art DP-SGD papers~\citep{de2022unlocking,li2022large} that base their accounting on Poisson subsampling but implement shuffling. They observe significantly higher empirical $\epsilon$ for the shuffling than provided by the privacy accounting that assumes Poisson subsampling, e.g., empirical $\epsilon=13.54$ for some training on the Places-365 dataset with theory indicating $\epsilon=7.53$.
\citet{chua2024private} provide a privacy accounting comparison between Poisson subsampling and shuffling and emphasize that the guarantees can be very different.

\citet{DBLP:conf/aistats/ChuaGHKKLMSZ25} offer an alternative to Poisson subsampling by randomly assigning each sample to precisely one batch per epoch and implementing an accountant for their method. \citet{DBLP:conf/iclr/Choquette-ChooG25} build on top of this with a Balls-in-bins minibatching using JAX and a similar technique to enforce fixed batch sizes. While these methods are complementary to ours, they modify the sampling scheme, and require a new computationally expensive accountant. We focus on correctly implementing Poisson subsampling and utilizing existing privacy accountants.\looseness-1

\section{Avoiding Re-compilation in JAX}\label{sec:masked-dp-sgd}

Using JAX for DP-SGD introduces complexities, particularly around Poisson subsampling which is crucial for privacy accounting. Implementing Poisson subsampling results in variable logical batch sizes that lead to variability in the size of the last physical batch which require JIT to recompile, leading to graph retracing which is costly and contributes to execution run variability \citep{chua2024private}. 

\subsection{Masked DP-SGD}
We propose an algorithm, called masked DP-SGD, that overcomes the issue of recompilation at the cost of computing slightly more gradients than the naive implementation while at the same time using proper Poisson subsampling and therefore ensuring the correct privacy budget. We execute the following sub-steps at every iteration and highlight the differences to the naive implementation (steps 2 and 4) (See also \cref{alg:dpsgd-jax}):

\begin{enumerate}[leftmargin=*,noitemsep,topsep=0pt]
    \item We sample a logical size using Poisson sampling.
    \item \emph{We round up the number of samples for which we compute per-sample gradients so that it is divisible by the physical batch size without remainder.}
    \item We compute the per-sample gradients.
    \item \emph{We mask out gradients so that the per-sample gradients used for the update are the actual Poisson subsampled ones, ensuring compliance with the Poisson subsampling accounting.}\looseness-1
\end{enumerate}

In order to improve the readability of our pseudo-algorithm in \cref{alg:dpsgd-jax}, we further
decompose Poisson subsampling into two steps: (1) draw a batch size from 
$b \sim \mathrm{Binom}(N, q)$ and (2) sample the subset of size $b$ from the full data
using without replacement sampling (WOR). The following Lemma shows that the decomposed procedure is equivalent to Poisson subsampling.\looseness-1

\begin{lemma}\label{lemma:poisson}
    Poisson subsampling from a data set $D$ of size $N$ with subsampling rate $q$ is 
    identical to the following procedure
    \begin{enumerate}
        \item draw $b \sim \mathrm{Binom}(N, q)$
        \item return a batch $B \sim \mathrm{WOR}(D, b)$.
    \end{enumerate}
\end{lemma}

We provide the proof in \cref{sec:poisson_proof}.

\begin{figure}[!h]
\begin{algorithmic}[1]
   \STATE {\bfseries Input:} Training data points $D = \left\{x_{1},\dots,x_{N}\right\}$, loss function $\mathcal{L}(\theta) = \frac{1}{N}\sum_i \mathcal{L}(\theta,x_i)$
   \STATE {\bfseries Parameters:} learning rate $\eta_t$, noise scale $\sigma$, gradient norm bound $C$, number of steps $T$, expected logical batch size $L$, physical batch size $p$.
    \STATE {\bfseries Start} 
   
   \FOR{$t \in [T]$}

   \STATE \textcolor{blue}{// {Sample the true batch size
   \STATE $b \sim \text{Binomial}\left(N, \frac{L}{N}\right)$} 
   \STATE // Find minimum $k \in \mathbb{N}$ such that $p \cdot k \geq b$ 
   \STATE $\color{blue}b_{+} \gets p \cdot k$
   \STATE // {Sample a batch of size \textcolor{blue}{$b_+$} without replacement}
   \STATE $B \gets \mathrm{WOR}(D, b_+)$}
   \STATE // {Divide $B$ into physical batches of size $p$}.
   \STATE $P \gets \{B_1,\ldots,B_k\}$
   \STATE // \textcolor{blue}{Create masks so that $\sum_{i=1}^{b_{+}} M_i = b$} 
   \STATE $\color{blue}M \gets \{\underbrace{1,1,\ldots,1}_{b},\underbrace{0,0,\dots,0}_{b_+-b}\}$ 
   \STATE $\theta_{acc} \gets \mathbf{0}$ 
   \FOR{$s \in P$}
   \FOR{$x_i \in s$}
   \STATE // {Compute gradient}
   \STATE $\textbf{g}_{t}(x_{i}) \leftarrow \nabla_{\theta_{t}} \mathcal{L}(\theta_{t},x_{i})$ 
   \STATE // {Clip gradient \textcolor{blue}{and mask}}
   \STATE $\overline{\mbox{\textbf{g}}}_t (x_{i}) \leftarrow \color{blue} M_{i+(s-1)*p} \cdot \color{black}\textbf{g}_t / \text{max} (1,\frac{\|\textbf{g}_t (x_i) \|_2}{C})  $ 
   \ENDFOR
   \STATE // {Accumulate gradient}
   \STATE $\theta_{acc}  \leftarrow \theta_{acc}+ \sum_{i} \overline{\mbox{\textbf{g}}}_t (x_{i})$ 
   \ENDFOR
   \STATE // {Add noise}
   \STATE $\widetilde{\textbf{g}}_{t} \leftarrow \frac{1}{L}(\theta_{acc} + \mathcal{N}(0,\sigma^{2}C^{2}\textbf{I}))$  
   \STATE // {Take step}
   \STATE $\theta_{t+1} \leftarrow 
   \text{OptimizerUpdate(}\theta_{t}, \widetilde{\textbf{g}}_{t}$)
   \ENDFOR
   \STATE{\bfseries Return} Learned parameters $\theta_T$ and the privacy cost from a privacy accountant.
\end{algorithmic}
   \caption{Our prosed algorithm Masked DP-SGD with differences to the default virtual batching algorithm in \cref{alg:dpsgd} highlighted in \textcolor{blue}{blue}.}
   \label{alg:dpsgd-jax}
\end{figure}

\subsection{Computational Cost}\label{sec:masked_jax_compute_cost}
We will analyse the computational cost of our proposed algorithm and compare to a recent work by \citet{chua2024scalable}. 

\textbf{Masked DP-SGD}
Our proposed algorithm rounds the logical batch size up to the closest larger integer divisible by the physical batch size to avoid recompiling. Hence, for any sampled logical batch size $b$, the difference between $b$ and the upscaled batch size $b_+$ will be in $\{0, \ldots, p-1\}$ for a physical batch size $p$.

Denoting the excess batch size with $\Delta_p(b)$, we can write
\begin{align}
    \mathbb{E}[b_+] = \mathbb{E}[b + \Delta_p(b)].
\end{align}
Now, we can form a simple upper bound for the expected relative increase of batch size given that $\mathbb{E}[b] = L$ as
\begin{align}
    \mathbb{E}[b_{+}] / \mathbb{E}[b] \leq 1 + (p-1)/L.
\end{align}

When working large number of samples and non-negligible sampling probabilities, the excess cost due to upscaling the batch size will be modest for feasible physical batch sizes. For example, in our experiments the expected batch size of the Poisson subsampling was $L=\numprint{25000}$, whereas the physical batch sizes extended up to 
$p=64$. The expected relative increase in computed gradients would be $0.252\%$.

\textbf{\citet{chua2024scalable}}
A recent work by \citet{chua2024scalable} proposed an alternative implementation for
JAX compilable implementation of Poisson subsampled DP-SGD. In their approach the
logical batch sizes are sampled from a truncated Binomial distribution. This affects the 
privacy guarantees of the models, and therefore they need to compensate the truncated
sampling by increasing the noise std. for DP-SGD. They suggest an approach for 
computing the truncation bound $b_+$ as 
\begin{align}
    b_+ = \inf_B \{ B \in [N] \mid \Psi(N, L, B) \cdot T \cdot (1+e^\epsilon) \leq \tau \delta\}
\end{align}
where $\Psi(N, L, B)$ denotes the survival function ($1-\texttt{cdf}$) of 
$\text{Binom}(N, L/N)$ at $B$ and $T$ are the number of steps. The parameter
$\tau$ effectively scales the size of the tails and is used to calibrate the
noise std by selecting $\sigma$ such that the hockey-stick divergence between
the Poisson subsampled Gaussian mechanisms is bound by $(1-\tau)\delta$.

\citet{chua2024scalable} choose $\tau=10^{-5}$, which keeps the noise std. increase
very small. In their implementation, the gradients are computed for $b_+$ randomly selected samples, after which the final samples are chosen
according to the batch size $b$ sampled from the truncated Binomial. Hence the
computational excess over regular Poisson subsampling (\cref{alg:dpsgd}) becomes $b_+-b$. 

\begin{figure}[!h]
    \centering
    \includegraphics{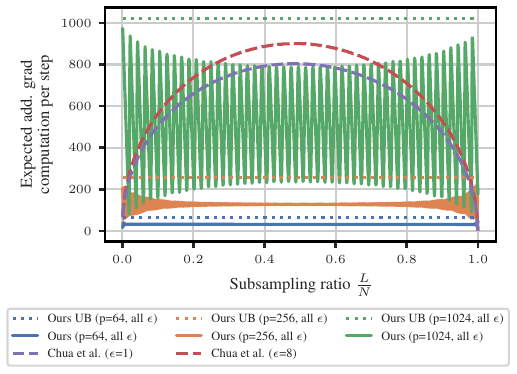}
    \caption{Expected additionally computed gradients per step of our proposed method and \citet{chua2024scalable}. We assume that the dataset size is $N=\numprint{50000}$ and simulate with $\epsilon \in \{1,8\}$ at $\delta=10^{-5}$ with 40 epochs for all $\nicefrac{L}{N}$, which means that the the number of iterations $T=40\times \nicefrac{N}{L}$. We set $\tau = 10^{-5}$ as done by \citet{chua2024scalable}. We also plot the simple upper bound that we computed a the beginning of the subsection with dotted lines. We observe that our method requires significantly fewer additional gradients for $p\leq256$.}
    \label{fig:computational_cost_sim}
\end{figure}

\textbf{Simulation}
Next, we will numerically evaluate the expected number of excess gradients $\mathbb{E}[b_+ - b]$ for both our and \citet{chua2024scalable} methods. For our method, we simply compute the excess for all batch sizes $b \sim \mathrm{Binom}(N, \nicefrac{L}{N})$ and evaluate $\sum_{i=0}^N \Pr(b = i) (i_+ - i)$. For truncated Binomial sampling by \citet{chua2024scalable}, we similarly compute the excess for all the batch sizes in the truncated domain, and evaluate the expectation.

In \cref{fig:computational_cost_sim}, we simulate the expected additional gradient computations per step $\mathbb{E}[b_+ - b]$ for both our method and \citet{chua2024scalable}.
We observe that our method has a significant smaller computational excess for most $\frac{L}{N}$ when the physical batch size $p\leq256$ which is the case when the model is large in relation to the VRAM of the used GPU (see \cref{sec:dpcost}). The difference is particularly large when the subsampling ratio $\frac{L}{N} \in [0.2, 0.8]$ which is reasonable used regime for subsampling ratios for training with DP-SGD~\citep{HowToDP-fy,räisä2024subsampling}. 

For the simulation we assume that the dataset size is $N=50000$ and we vary the subsampling ratio $q = [0.01, 1.0]$.
For our method we simulate with physical batch sizes $p \in \{64,256,1024\}$.
While our method is independent of the privacy budget, \citet{chua2024scalable} require specifying it. We simulate with $\epsilon \in \{1,8\}$ at $\delta=10^{-5}$ and simulate with 40 epochs for all q which means that the the number of iterations $T=\nicefrac{40N}{L}$. We set $\tau = 10^{-5}$ as done by \citet{chua2024scalable}.

Looking at the green line of \cref{fig:computational_cost_sim} where $p=1024$ we observe that the expected additional gradient computations per step $\mathbb{E}[b_+ - b]$ differ drastically for slightly different subsampling ratios $\frac{L}{N}$. For example, with $\frac{L}{N}=0.5$ the excess is $\mathbb{E}[b_+ - b]=599.92$ whereas with $\frac{L}{N}=0.51$ it reduces to  $\mathbb{E}[b_+ - b]=288.73$. There is an optimal physical batch size $p$ for every subsamping ratio $\frac{L}{N}$. 

\begin{figure}
    \centering
\includegraphics[width=\linewidth]{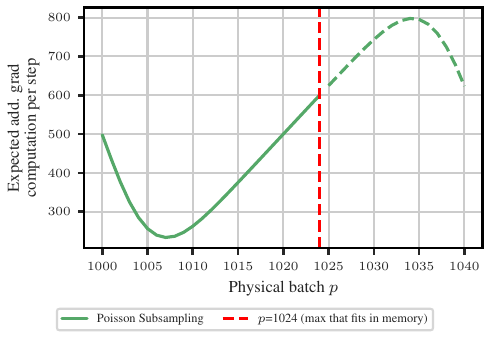}
    \caption{Expected additionally computed gradients per step of our proposed method. This is an illustration of an imaginary scenario where we assume that the dataset size is $N=50000$ and simulate with $\frac{L}{N}=0.5$. Choosing the maximum physical batch size $p=1024$ is not optimal in terms of expected additionally computed gradients per step.}
    \label{fig:sim_vary_p}
\end{figure}

In \cref{fig:sim_vary_p}, we illustrate this observation by plotting the excess $\mathbb{E}[b_+ - b]$  as a function of physical batch size $p$ for example choices of $N=50000$ and $\frac{L}{N}=0.5$. Imagine that the GPU VRAM can fit a physical batch size of $p=1024$, which is denoted with the dashed red line. In terms of expected excess computed gradients $\mathbb{E}[b_+ - b]$ it would be better to use a smaller physical batch size, e.g., $p=1007$.

\section{Experiment Overview}

\textbf{PyTorch implementations} We benchmark a native PyTorch~\citep{PyTorchCitation} implementation with PyTorch-based libraries Opacus~\citep{OpacusPaper} (details on gradsampling below), PrivateVision (PV)~\citep{PrivateVision}, and FastDP (BK) \citep{FastDPbu2022differentially}, see \cref{tab:algo}. At submission time ghost clipping in Opacus was still undergoing changes and was unstable in our experiments.

\textbf{JAX implementations} We benchmark two JAX implementations. Our method Masked DP-SGD and a native JAX \citep{jaxgithub2018} implementation that clips the per-sample gradients with Optax \citep{deepmindjaxoptax2020} without utilizing any further optimization. This naive implementation in JAX is prone to recompilation due to changing tensor sizes caused by the Poisson subsampling.

\textbf{Implementation of Poisson sampling}
Opacus samples the logical batches using Poisson sampling and then divides them into physical batches using their \texttt{BatchMemoryManager} class. The other PyTorch implementations considered in our experiments do not support virtual batching out-of-the-box. To make a fair comparison between all methods, we implemented Poisson subsampling in the same way as Opacus for all frameworks and adapted the \texttt{BatchMemoryManager} to support them. Thus, all experiments are seeded to ensure the same logical batch sizes.

\textbf{Metrics} We compare the throughput, defined as how many samples can be processed per second during training, and the maximum physical batch size that can fit in  memory. We determine the maximum physical batch size before running out of memory (OOM) using binary search. In \cref{subsec:nlp}, throughput will be defined as the number of tokens processed per second.

\textbf{Dataset} We benchmark with the CIFAR100 \citep{KrizhevskyApril82009} resized to 224x224. In the appendix, we also evaluate additional datasets and present their benchmark results. We benchmark the IMDB \citep{IMDB} dataset for sentiment analysis in \cref{subsec:nlp}.

\textbf{Models} We train two families of CV models: Vision Transformer (ViT)~\citep{dosovitskiy2021an}\footnote{Taken from \url{https://huggingface.co/timm/vit_base_patch16_224.orig_in21k}}and  ResNet~\citep{ResNet}\footnote{Taken from \url{https://github.com/google-research/big_transfer}} (See \cref{tab:Size_Models_Tables}). Both are pre-trained on ImageNet-21k~\citep{ILSVRC15}. 
For our NLP evaluation, we fine-tune bert-base-uncased model, with 110M parameters \citep{Bert_model}. It is pre-trained on BookCorpus \citep{moviebook} and English Wikipedia.

\begin{table}[h]
    \caption{Number of parameters (millions) for used models.}
    \label{tab:Size_Models_Tables}
    \begin{center}
        \begin{small}
                \begin{tabular}{rr|rr}
                     \toprule
                     \multicolumn{2}{c|}{\textbf{Vision Transformer (ViT)}} & \multicolumn{2}{c}{\textbf{ ResNet}}\\
                     \textbf{Type} & \textbf{\# Params} & \textbf{Type} & \textbf{\# Params} \\
                     \midrule
                     Tiny &  5.7 M & $50{\times}1$ &  23.7 M \\
                     Small & 22.1 M & $101{\times}1$ & 42.7 M \\
                     Base & 86.6 M &$50{\times}3$ & 211.8 M\\
                     Large & 304.3 M &$101{\times}3$ & 382.4 M \\
                     Huge & 630.8 M &$152{\times}4$ & 929.2 M\\
                    \bottomrule
                \end{tabular}
        \end{small}
    \end{center}
\end{table}

\textbf{Parameterization} While parameter-efficient fine-tuning of some parts of the model has been shown to be effective under DP~\citep{yu2022differentially,Tobaben2023Efficacy}, our work focuses on the computational efficiency of DP-SGD and thus we consider the worst-case scenario of fine-tuning all parameters of the model. Furthermore, any training from scratch requires training all parameters.

\textbf{Hyperparameters} We train for four optimization steps with a sampling rate of $0.5$ (expected batch size of 25000), which allows us to quickly test the experiments with a realistic high batch size~\citep{HowToDP-fy,räisä2024subsampling}. The rest of the DP-SGD hyperparameters are discussed in \cref{sec:app-hyperparameters}. We do not focus on finding the best possible utility, which requires training for many more epochs (See \cref{tab:acc_table} for the accuracy after training for four steps).\looseness-1

\textbf{Environment specifications} We use two GPU architectures: NVIDIA V100 (32 GB VRAM) and A100 (40 GB VRAM) with identical Python environments. Each node contains four GPUs. We use 16 CPU workers for data loading. In the distributed case of more than one GPU, only one worked per device is used.\looseness-1

\textbf{Source code}
We provide the code for reproducing the experiments and our implementation as a library\footnote{\url{https://github.com/DPBayes/Towards-Efficient-Scalable-Training-DP-DL}}.

\textbf{Grad sample modes in Opacus}\label{app:gradsample}
Opacus supports multiple different gradient sampling methods as indicated in the documentation\footnote{\url{https://github.com/pytorch/opacus/tree/61ae0ea4fb37a835e93040b5de19e8dfcd465a07/opacus/grad_sample}}. In our original experiments we used the grad\_sample mode \texttt{hooks} that is the default. This will use custom opacus modules when they are defined for that layer and functorch as a fallback. We tried out different methods listed in the documentation for both ResNet and ViT models:

\begin{itemize}[leftmargin=*]
    \item \texttt{functorch}: We forced opacus to use functorch but did not observe any significant speed differences to using \texttt{hooks}. This is in line with the opacus documentation which writes that the speed is $0-50\%$ slower than \texttt{hooks}.
    \item \texttt{ExpandedWeigths}: We tried this approach but ran into runtime errors. Interestingly, when looking through the issues others have reported issues\footnote{\url{https://github.com/pytorch/opacus/issues/464 }}\footnote{\url{https://github.com/pytorch/opacus/issues/584}} but it seems to be more a PyTorch problem and has not been addressed for years. According to the documentation \texttt{ExpandedWeigths} is still in beta status.
    \item \texttt{GhostClipping}: Note that this method only works for ViT as described in \cref{sec:exp_optimized_dp-sgd}. At first we did not manage to decrease the loss with this implementation due to the implementation in opacus being unstable. After some fixes, the correct accuracy is achieved but we noticed that the speed-ups are not significant, and even lower than flat clipping. Therefore, we decided to not include them, as it is still in development. When ready, we expect a similar speed-up to the observed in our experiments in \cref{sec:exp_optimized_dp-sgd} as the underlying algorithm is the same.
\end{itemize}

\section{What is the Computational Cost of DP in Deep Learning \label{sec:dpcost}}

We quantify the computational cost of deploying DP training by comparing the throughputs and maximum physical batch sizes between the non-private training with PyTorch and private training with Opacus, the most widely used DP-SGD implementation. Additionally, we identify the reasons for the higher computational cost of DP-SGD through profiling.

\subsection{Throughput and Maximum Batch Size Comparison}\label{subsec:cv_comp_thr}

\begin{figure*}[tb]
    \begin{center}
        \includegraphics[width=0.49\textwidth]{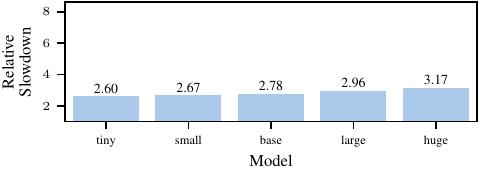}
        \includegraphics[width=0.49\textwidth]{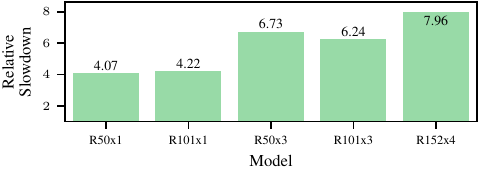}
        \caption{Relative slowdown (left: Vision Transformer, right: ResNet) in mean throughputs between Opacus per-example clipping and the non-private baseline (A100 GPU). The relative slowdown is calculated as the ratio of private-throughput to non-private-throughput. A lower value indicates a better performance, closer to 1 indicates that Opacus is as fast as non-private training. This highlights the computational cost associated with private training.}
        \label{fig:relative_diff_model_sizes}
    \end{center}
\end{figure*}

\begin{figure*}[tb]
	\begin{center}
		\includegraphics[width=0.49\textwidth]{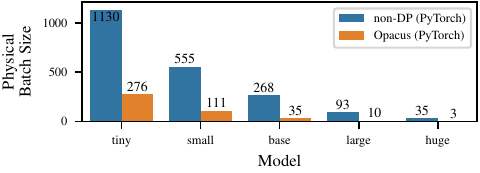}
		\includegraphics[width=0.49\textwidth]{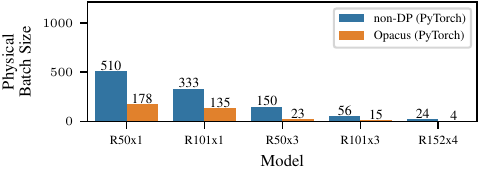}
		\caption{Maximum achievable physical batch size by the different model sizes (left: Vision Transformer, right: ResNet) on A100 GPU (40 GB) before they reach Out Of Memory (OOM) Error. The model sizes grow from left to right (Refer to \cref{tab:Size_Models_Tables} for number of parameters).}
		\label{fig:max_bs_model_sizes}
	\end{center}
\end{figure*}

We compare relative throughput (\cref{fig:relative_diff_model_sizes}) and the maximum physical batch size (\cref{fig:max_bs_model_sizes}) between DP-SGD (Opacus) and non-private training with PyTorch. The main metric of interest is the throughput as it quantifies the training speed, but the maximum physical batch size becomes important when training models that are too large to fit even one example at a time. For both metrics, DP-SGD becomes more expensive with larger models, but the detailed trends differ.

\textbf{Vision Transformer} The throughput difference between Opacus and the non-private baseline with PyTorch (see left of \cref{fig:relative_diff_model_sizes}) grows steadily as a function of model size, which is interesting considering how big the relative difference in the maximum physical batch size (right of \cref{fig:relative_diff_model_sizes}) is: 
the throughput ranges from a relative difference of $\times 2.6$ for the smallest model to $\times 3.17$ for the largest model while the maximum physical batch size has a relative difference of around $\times 4$ for the smallest model and $\times 11$ for the largest model.

\textbf{ResNets}
On \cref{fig:relative_diff_model_sizes}, we observe a more irregular throughput and relative slowdown for the ResNets models size as their size grows. The contrast in \cref{fig:relative_diff_model_sizes} between ViT and ResNet models is due to the architecture and types of layers. The parameter space grows as the width factor (see \cref{tab:Size_Models_Tables}) for the ResNets, so the $\times 3$ makes the neural network wider by a factor of three. Based on our results, the width of the layers affects throughput much more than the depth of the network. ResNet models with the same width and different depths exhibit comparable throughput, but increasing the width will make the model in the private setting much slower and reduce the maximum batch size significantly.

\begin{figure}[t]
  \begin{center}
      
      \centerline{\includegraphics[width=\columnwidth]{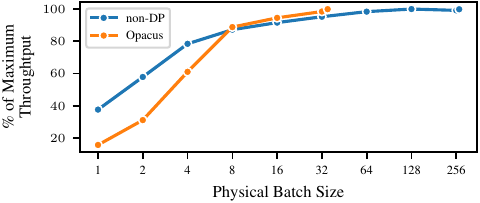}}
      \caption{The relative difference with the throughput at the maximum batch size for the ViT base model on A100.}
      \label{fig:relative_thr_vs_bs}
  \end{center}
\end{figure}

\textbf{How much does finding the maximum physical batch size matter?}
In \cref{fig:relative_thr_vs_bs}, we display the relative throughput as a percentage by dividing the throughput at a particular physical batch size by the maximum achievable throughput. We see that as the physical batch size increases, the throughput will grow as expected, but there is no significant further improvement at some point. Practitioners may estimate the optimal batch size based on available memory and performance trade-offs. Using the maximum physical batch size is not crucial, but a large enough value is sufficient. Typically, the throughput of smaller batches is limited by data loading speeds, but computation becomes the limiting factor as batch size increases.

\begin{table}[t]
    \caption{Average processing time in milliseconds for each section of the algorithm. We are comparing the non-private and Opacus clipping on A100 (same physical batch size). We profile the time using NVIDIA Nsight Systems. All measurements include synchronization time, which is needed for the profiling, but adds additional time that is not part of the normal execution.\looseness-1}
    \label{tab:nsys}
    \begin{center}
        \begin{small}
            \begin{sc}
            \begin{adjustbox}{max width=\linewidth}
                \begin{tabular}{lcc}
                    \toprule
                    Section & \multirow{2}{2cm}{\centering non-DP (PyTorch)} & \multirow{2}{2cm}{\centering Opacus (PyTorch)}  \\
                    & & \\
                    \midrule
                    Forward & 81.14 & 101.53 \\
                    Backward & 163.85 & 681.48 \\
                    Clip \& Accumulate & 0 & 26.76\\
                    Optimizer Step & 38.17 & 99.65 \\
                    \bottomrule
                    \end{tabular}
            \end{adjustbox}
            \end{sc}
        \end{small}
    \end{center}
\end{table}

\subsection{Reasons for the Increase in Computational Cost}\label{sec:profiling}
Providing a detailed breakdown of low-level operations associated with DP is challenging. However, using GPU profiling tool NVIDIA Nsight System, we can identify three aspects which constitute the majority of DP overheads. Firstly, due to its larger memory footprint, DP-SGD is limited to consume smaller physical batches than its non-private counterpart. This results in a larger amount of smaller low-level kernel calls, which leads to slightly lower utilization of the GPU. Even kernel launch overheads can become a notable factor for a slowdown at very small batch sizes. Secondly, the computation of per-example gradients introduces significant overhead in the backward pass as it cannot be parallelized as in batched gradient computation. This is the most prominent cause of the total overhead. Finally, an additional DP-optimizer step that clips and accumulates the per example gradients, which is not present in the non-DP algorithm, must be taken after each physical batch (see \cref{tab:nsys}).\looseness-1

\section{Decreasing the Computational Cost}

This section analyzes the different strategies for training with DP-SGD more efficiently. We evaluate both algorithmic and hardware optimizations and their combinations.

\subsection{Efficient Gradient Clipping Algorithms}\label{sec:exp_optimized_dp-sgd}
First, we evaluate more efficient gradient clipping implementations that have been described in \cref{subsec:optim} using the Vision Transformer base model. We chose it as our benchmark model because the middle model size is large enough to evaluate the advantages of the optimized gradient clipping algorithms but does not require excessive amount of time to train. The non-Opacus implementations do not support the ResNet due to their custom weight standardization layer.

\begin{figure}
    \centering
    \includegraphics[width=\linewidth]{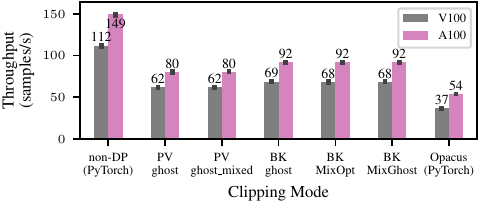}
    \caption{Throughput using the maximum batch size for each clipping algorithm. It compares the executions for both V100 and A100, for the ViT Base model.}
    \label{fig:thrs_gpus_comp}
\end{figure}

\textbf{Throughput Comparison}
\cref{fig:thrs_gpus_comp} displays the throughput for each clipping algorithm for each tested GPU. Moving from a V100 to an A100 GPU increases throughput by $\times 1.3$ times on average over all clipping methods. The one that benefited the most is the per-example clipping by Opacus with a $\times 1.46$ improvement in throughput. This is because of Opacus-specific optimizations. Their implementation is optimized to vectorize the virtual batches and get the most out of the processing unit to compensate for the per-example clipping. We base our virtual batching module on Opacus, which may have further contributed to the advantage seen for Opacus. The other implementations showed benefits similar to those of non-private training. For both GPUs, the clipping optimizations consistently maintained their relative throughput difference to their non private baseline (see \cref{fig:ThrvsBS_JAX_PyTorch} upper). Private Vision gets closer to the non-private baseline physical batch size, but Book Keeping is closer to its throughput with a smaller physical batch size (see \cref{fig:precision_comp} right).

Without sacrificing utility (see \cref{tab:acc_table}), these optimizations offer an alternative to the original per-example clipping algorithm. Although Book Keeping has a slightly better throughput, the margin is narrow, making Private Vision and FastDP viable options as ghost clipping implementations. The difference between the two algorithms is the second backward pass over the neural network. The Book Keeping trick avoids this second backward pass, resulting in higher throughput at a small memory cost.

Mixed ghost clipping does not yield any improvement because it determines whether to apply ghost or per-example clipping, based on the size of the inputs and the parameter space. For large dimensions, ghost clipping will be more expensive \citep{PrivateVision}. In ViT models, the dimensions change less than in a convolutional network. Therefore, despite continually evaluating which method to apply, it consistently defaults to ghost clipping. Conversely, mix optimization applied to a ResNet model should outperform ghost clipping since it is optimized for convolutional layers. This could not be tested on ResNet models due to incompatibilities with Private Vision and FastDP, preventing an assessment of mixed optimization methods.\looseness-1

\textbf{Maximum physical batch size}
\cref{tab:max_bs_table} compares the maximum physical batch size for both available GPUs. The maximum physical batch size is larger for the optimizations of DP-SGD than for Opacus because they do not require per-example gradients. Consequently, these optimizations enable training much larger models without exhausting memory. The maximum physical batch size using the Private Vision library is the closest to the non-private baseline. Generally, the methods are consistent within implementations, with Private Vision and FastDP achieving the same maximum physical batch size regardless of the clipping mode. As expected, the A100 consistently attains higher maximum physical batch sizes than the V100 due to its larger VRAM.

\begin{table}
    \caption{Maximum physical batch size reachable for each clipping method and GPU using for the ViT base model.}
    \label{tab:max_bs_table}
    \begin{center}
        \begin{small}
            \begin{sc}
                \begin{tabular}{ l|c|c } 
                    \toprule
                    \textbf{Clipping Mode} & \textbf{V100} & \textbf{A100} \\
                    & (32GB) & (40GB)\\
                    \midrule
                    Non private baseline & 216 & 268 \\
                    Per-example (Opacus) & 28 & 35 \\ 
                    Ghost (Private Vision) & 203 & 257 \\
                    Mix Ghost (Private Vision) & 203 & 257 \\
                    BK Ghost (FastDP) & 189 & 209 \\
                    BK Mix Ghost (FastDP) & 189 & 209 \\
                    BK Mix Opt (FastDP) & 189 & 209 \\
                    \bottomrule
                \end{tabular}
            \end{sc}
        \end{small}
    \end{center}
\end{table}

\subsection{Lower Precision}\label{subsec:lower}

We consider using lower precision to accelerate computation. We evaluate the use of TensorFloat-32 (TF32) for training. TF32 has 10 bits for precision, with eight range bits, giving it the same range but less precision than 32-bit single precision floats (FP32)~\citep{Nvidiatf32}. Using lower precision can have benefits exactly where DP training struggles: it requires less memory and uses fewer bits to represent the data, and its operations are optimized for GPU, making them much faster \citep{MixPrecisionCUDA}. It is special math mode introduced for the A100 GPU and unavailable for the V100, so we compared training on the A100 with and without TF32.\looseness-1

\begin{figure}[ht]
    \begin{center}
        \centerline{\includegraphics[width=\linewidth]{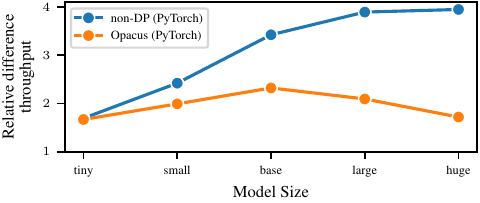}}
        \caption{Relative difference in mean throughput between TF32 and FP32 Training for ViT Models. The throughput ratio is calculated as TF32 divided by FP32.}
        \label{fig:size_mahti_tf32}
    \end{center}
\end{figure}

\textbf{Experimental results} In \cref{fig:size_mahti_tf32}, we display the relative difference between mean throughput using TF32 and FP32. For non-private training, throughput increases with model size. For private training throughput increases for the smaller models, but it goes down again as the model size grows after the base size. Models that are too small do not gain much from TF32, and the larger ones have too small maximum physical batch size to benefit.
The maximum physical batch size is unaffected by the precision, which is expected since TF32 is a tensor core operation and not a data type \citep{tf32discussion}.

\textbf{Model size TF32 speedups} The speedup observed in \cref{fig:size_mahti_tf32} peaks at the "base" model. We believe that the reasons are the following: Speed-ups resulting from TF32 can significantly vary on per case basis as “all storage in memory and other operations remain completely in FP32, only convolutions and matrix-multiplications convert their inputs to TF32 right before multiplication.” \citep{tf32discussion}. Until now, TF32 precision benchmarks have been limited to non-DP applications which was one of the reasons we wanted to discuss our observations in a DP context. It appears the effectiveness of TF32 arithmetic peaks at “base” configuration. This due to a mix of reasons which are difficult to quantify exactly. Firstly, it is likely that matrix multiplication kernel dominance peaks at this configuration i.e. we have the most parameters whilst the batch size dimension also remains sufficiently large. With large and huge model variants the parameter count still increases but at the cost  having very small batch dimension of 10 and 3, respectively. Secondly, we observe similar trend in the left side of \cref{fig:relative_diff_model_sizes} where the discrepancy between DP and non-DP grows as model size gets bigger. This suggests that the dominance of DP operations also grows with the model size. None of the DP-operations are cast as matrix-multiplications and hence won't benefit from TF32.

\textbf{Concerns regarding TF32 under DP}
There are two concerns with using lower precision in DP deep learning: its effects on utility and privacy. Lower precision may affect utility, as it is less precise. We did not find a significant decay in the accuracy of the models compared to the models with FP32; it differs by decimal points at the $\times 10^{-6}$ precision (See \cref{tab:acc_table,tab:accuracy_comparison}).
Regarding privacy, all floating point implementations provide imperfect implementations of real-valued mechanisms, that might introduce additional privacy vulnerabilities \citep{BitsDP}.
Lower precision may exacerbate this issue.
Discrete mechanisms \citep[e.g.][]{discrete_gaussian, skellam} avoid these theoretical challenges, but are often less convenient and may reduce utility, especially in low precision settings.
The efficiency of different discrete mechanisms in TF32 is an interesting topic of further research.\looseness-1

\begin{table}[ht]
    \caption{Mean accuracy for CIFAR-100 test set for each clipping mode for the ViT models on A100 after training for two epochs. All use the ViT hyperparameters from \cref{tab:hypers}. While this work does not focus on the model's utility, having their results still allows us to compare them. 
    The use of TF32 as a lower precision mode does not affect the model's utility.}
    \label{tab:acc_table}
    \begin{center}
        \begin{small}
            \begin{sc}
                \begin{tabular}{lcc}
                    \toprule
                    Clipping Mode & Test Accuracy \\
                    \midrule
                    Opacus & 0.8223\\
                    Opacus/TF32 &  0.8225 \\
                    JAX naive &  0.8146 \\
                    Masked DP-SGD & 0.8224 \\
                    PV-ghost & 0.822 \\
                    BK-ghost & 0.822 \\
                    \bottomrule
                    \end{tabular}
            \end{sc}
        \end{small}
    \end{center}  
\end{table}
\vspace{-0.1in}

\section{Comparison of JAX Implementations}\label{sec:jax}

We provide a performance comparison for the use of PyTorch and JAX across two domains, computer vision and natural language processing, both for a classification task. \cref{subsec:cv} compares the performance of a naive non-private JAX, a naive JAX, and our proposed masked DP-SGD method with all other DP-SGD frameworks (all based on PyTorch). The utility is the same as in PyTorch (See \cref{tab:acc_table}). \cref{subsec:nlp} compares our proposed method with Opacus. To provide a fair comparison, we implemented non-private and native DP JAX training using the same virtual batching as PyTorch. Note that JAX defaults to TF32 when available, and FP32 needs to be explicitly forced.\looseness-1

\subsection{Computer Vision} \label{subsec:cv}

\begin{figure*}[ht]
    \begin{center}
    \centerline{\includegraphics[width=\linewidth]{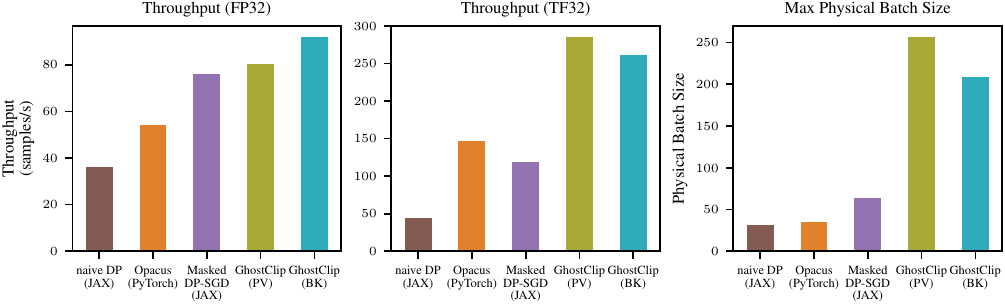}}
        \caption{Throughput comparison across precision modes, for the ViT Base model, trained on A100 GPU.
        Using a lower precision should increase memory capacity and speed-up the sample processing. The results confirm that throughput is enhanced with lower precision. However, the physical batch size remained constant across precision modes.}
        \label{fig:precision_comp}
    \end{center}
\end{figure*}

\begin{figure}[!ht]
\begin{center}
  \includegraphics[width=\columnwidth]{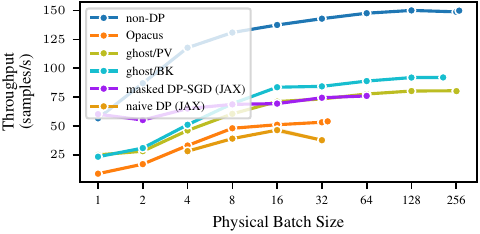}
    \hfill
   \includegraphics[width=\columnwidth]{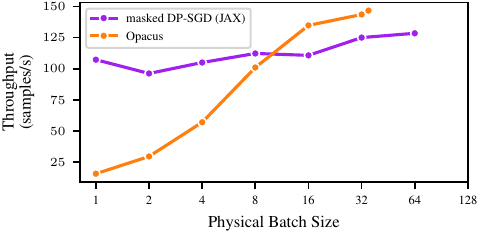}
   \caption{Comparison of the throughput as a function of the physical batch size between the JAX and PyTorch clipping algorithms on A100 GPU (upper: FP32, lower: TF32). In the lower subplot, we compare our method against Opacus on TF32 precision. The analysis excludes the Mix algorithms, due to their equivalent performance in ViTs. 
   }
   \label{fig:ThrvsBS_JAX_PyTorch}%
\end{center}
\end{figure}

\textbf{Throughput comparison (FP32)} In \cref{fig:precision_comp} (left), we compare the throughput using FP32. The naive DP-SGD JAX is the slowest implementation due to the JAX recompilation. Our proposed method masked DP-SGD outperforms Opacus and nearly matches the performance of PV Ghost Clipping despite not utilizing any optimizations regarding clipping. The masked DP-SGD exhibits higher throughput compared to other JAX implementations. This is primarily because the entire logical batch is accommodated in CPU memory, allowing it to be split into static sizes. Consequently, the compilation time is elevated for the first logical batch; however, subsequent iterations benefit from increased speed as recompilation is unnecessary. In \cref{fig:precision_comp} (middle) we compare throughput using TF32 and for this precision, the results indicate that masked DP-SGD performs comparably to Opacus in terms of throughput. However, our method performs better on regimes with fewer samples (\cref{fig:ThrvsBS_JAX_PyTorch}) and allows for a larger physical batch size (\cref{fig:precision_comp} right).\looseness-1

\textbf{Compilation}
The compilation time must be taken into account, as DP-SGD implementations in PyTorch do not compile. We measure it as the duration to process the first batch, since the execution times for each batch show that the first batch takes much more time than the others, including the compilation time (see \cref{fig:compilation_jax}).
The compilation time increases with batch size.
For the private model, the compiled function is more complex than the non-private counterpart. It includes expanding the dimensions and clipping the gradients, while the non-private directly computes the gradient of the whole mini-batch.\looseness-1

Although compiling PyTorch is possible, we did not observe any significant speed improvements. Compiling the non-private model yielded minimal speed-up, but ultimately even lower when accounting for the compilation. PyTorch also recompiles after a batch size change, but reverts to predefined CUDA optimized operations. In the private setting, the compilation does not recognize Opacus hooks and continues the execution without compiling them (See \cref{fig:compilation_torch}). Leveraging the same kernels to support the private hooks and avoid the compilation would require massive engineering work of writing special kernels for each specific private case. On the other hand, JAX will compile the JIT functions in XLA, but it does not fall back to the kernels, making it more generalizable \citep{EnablingJIT2021}.\looseness-1

\textbf{Variability in experiments}
One difference between the two frameworks is the variability in the experiments. PyTorch runs are remarkably consistent, maintaining low variance, and yielding the same throughput result for a fixed seed. In contrast, naive JAX experiments trigger recompilation and show greater variability than those of PyTorch, likely due to its sensitivity to fluctuations in the HPC environment and accelerator stochasticity, as noted in \cref{fig:compilation_jax}. Additionally, JAX's asynchronous dispatch method complicates time benchmarking by issuing a promise rather than immediate results, thereby concealing Python overheads. For our Masked DP-SGD method, by fixing the batch sizes and avoiding recompilation, we achieve consistent execution times.\looseness-1

\subsection{NLP Experiments}\label{subsec:nlp}

\begin{figure}[ht]
    \begin{center}
        \centerline{\includegraphics[width=\columnwidth]{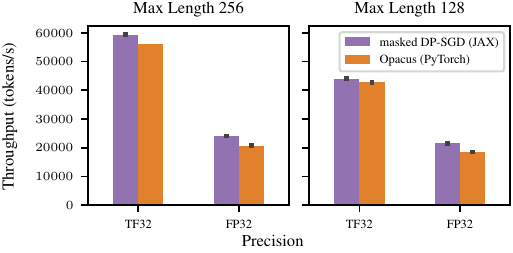}}
        \caption{Comparison of the throughput as a function of the precision and the max length, with a physical batch size of 64, for the bert-base-uncased model fine-tuned on the IMDB dataset on A100 GPU.}
        \label{fig:bert_exp}
    \end{center}
\end{figure}

For the NLP experiments, we fine-tune the BERT (bert-base-uncased) model on the IMDB dataset. For both frameworks, we freeze the embedding layer. Unlike computer vision tasks, where throughput is measured in images per second, we report tokens per second to better reflect the variable-length nature of text sequences. NLP introduces an additional computational constraint through the maximum sequence length hyperparameter. During tokenization, sequences are either truncated or padded to this maximum length; while tokenizing without length constraints preserves all information, it creates variable-sized tensors that are computationally inefficient for batched processing. \cref{fig:bert_exp} shows throughput comparisons for maximum lengths of 128 and 256 tokens. Our Masked DP-SGD implementation in JAX achieves a speedup of $\approx 1.15$ compared to Opacus for the FP32 precision.\looseness-1

\begin{figure}[htb]
    \begin{center}
        \centerline{\includegraphics[width=\columnwidth]{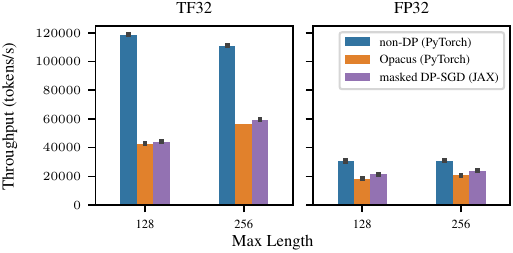}}
        \caption{Comparison of throughput as a function of max length and precision, with a physical batch size of 64, for the bert-base-uncased model fine-tuned on the IMDB dataset on A100 GPU, including the non-DP baseline training mode. The plots are divided by precision to illustrate that precision has a more significant impact on efficiency than the maximum length choice.}
        \label{fig:bert_exp_non}
    \end{center}
\end{figure}

\textbf{Precision comparison} Reducing precision from FP32 to TF32 provides substantial throughput improvements: approximately 2.6× faster processing for maximum length 256 and 2.1× faster for maximum length 128. But we can see the impact even more in the non-DP case, where it is 3.5x faster. Although longer sequences show greater absolute speedup gains, the relationship between maximum length and throughput follows a U-shaped curve. As the maximum length increases, more sequences require padding, resulting in processing of essentially empty tokens that reduce computational efficiency. The optimal maximum length is dataset-dependent and represents an important area for future research. Both algorithms benefit similarly from reduced precision, with NLP tasks showing greater relative improvements from precision reduction compared to computer vision tasks, as we see in \cref{fig:bert_exp_non}. We do not observe significant differences in accuracy regarding the precision and training optimization, as we see in \cref{tab:acc_table_bert}.\looseness-1

\section{Distributed Training}\label{sec:multgpu}

We will look at another angle for training deep learning with DP faster: increasing the computational resources enough to decrease the training time. This is relevant when training cost or resource constraints are less important than the time to train a new model.\looseness-1

We utilize V100 GPUs on HPC nodes that have 4 GPUs per node. The other experimental setting is identical to the one in \cref{sec:dpcost}. Results for utilizing up to 24 A100 GPUs can be found in \cref{fig:GPUs_wtf32_v2} in the Appendix. We focus on comparing the scaling behavior between the non-private baseline that uses PyTorch and the DP-SGD implementation using Opacus as both frameworks provide mature tooling for distributed training. 

\begin{figure}[!ht]
    \begin{center}
    \centerline{\includegraphics[width=\linewidth]{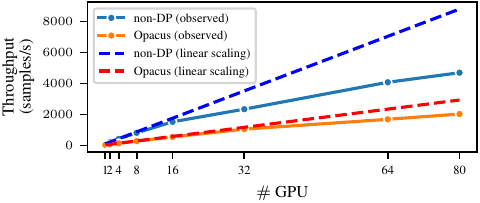}}
        \caption{Comparison between the throughput by scaling the number of GPUs for the non-private and Opacus training with the ViT base model on V100 GPUs. The dashed line is the ideal growth. }
        \label{fig:multinode_puhti}
    \end{center}
\end{figure}

\begin{figure}[!ht]
\begin{center}
  \includegraphics[width=\columnwidth]{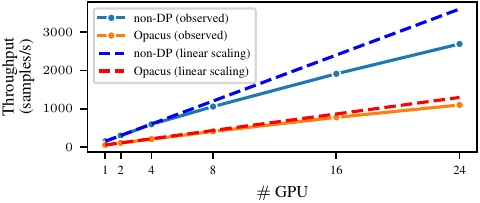}
    \hfill
   \includegraphics[width=\columnwidth]{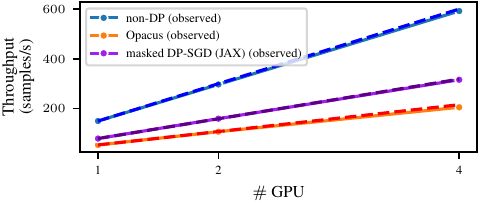}
   \caption{Comparison between the throughput by scaling the number of GPUs with more nodes for the non-private and Opacus training with the ViT base model on A100 GPUs. The dashed line is the ideal growth if it were linear. In the lower subplot we compare the previous methods with our masked DP-SGD implementation. It scales almost linearly, in a one node, 4-GPU setting. 
   }
   \label{fig:GPUs_wtf32_v2}%
\end{center}
\end{figure}

\cref{fig:multinode_puhti} shows the throughput increase as a function of number of GPUs. The throughput does not grow linearly and deviates from ideal linear scaling after using more than one node (i.e. more than 4 GPUs). While the communication inside the node is fast, the communication between nodes will always be slower. The bottleneck is the network bandwidth, and it prevents the throughput from scaling linearly. Notably, it affects the non-private training baseline much more, while the private scales close to optimal up to 32 GPUs. For the 80 GPUs, the private training achieves $69.2\%$ of the ideal linear speed-up, and the non-private training only achieves $53.3\%$. Private training scales better because it is slower and only sometimes saturates the network with updates.  

If we use Amdalh's law to compare the parallelism percentage for each case, we can see that in the private case, we achieve a $99.5 \%$ parallelism compared to a $98.9\%$ in the non-private case (See \cref{fig:amdahls_law_puhti_log}). We expect that the findings are not limited to PyTorch but are more general (see \cref{sec:discussion}).

\begin{table*}[!htbp]
\caption{A summary of the lessons learnt. The relative throughput/max physical batch size is in comparison to PyTorch non-DP (higher is better) on A100. For each optimization method and each model size, we divide it with the non-private counterpart.}\label{tab:lessons_learnt}
\arrayrulecolor{black!30}
\begin{center}
\begin{adjustbox}{max width=\linewidth}
\begin{tabular}
{|l|c|c|c|cc|l|}
\hline
\multirow{2}{*}{\textbf{Method}} & \multicolumn{2}{c|}{\textbf{Relative to non-DP (PyTorch FP32)}} & \textbf{Supports} & \multicolumn{2}{l|}{\textbf{Compilation}} & \multirow{2}{*}{\textbf{Section}} \\ \cline{5-6}
 & \textbf{Throughput ($\uparrow$)} & \textbf{Max Physical Batch Size ($\uparrow$)} & \textbf{all layers} & \multicolumn{1}{l|}{\textbf{Initial}} & \textbf{Re-}  &  \\ \hline
\textbf{Opacus} & 0.31-0.39 & 0.08-0.24 & \tikzcmark & \multicolumn{1}{l|}{-} & - & \cref{sec:dpcost} \\ \hline
\textbf{Efficient Gradient Clipping} & 0.49-0.54 & 0.88-0.95 & \tikzxmark & \multicolumn{1}{l|}{-} & - &  \cref{sec:exp_optimized_dp-sgd} \\ \hline
\textbf{Native JAX} & 0.39-0.59 & 0.23-0.43 & \tikzcmark & \multicolumn{1}{l|}{\tikzcmark} & \tikzcmark  & \cref{sec:jax} \\ \hline
\textbf{Masked DP-SGD (ours)} & 0.51-0.69 & 0.11-0.23 & \tikzcmark & \multicolumn{1}{l|}{\tikzcmark} & \tikzxmark & \cref{sec:jax} \\ \hline
\textbf{Masked DP-SGD + TF32} & 0.79-1.33 & 0.11-0.23 & \tikzcmark & \multicolumn{1}{l|}{\tikzcmark} & \tikzxmark  & \cref{sec:jax} \\ \hline
\textbf{Low Precision (Opacus+TF32)} & 0.54-0.84 & 0.08-0.24 & \tikzcmark & \multicolumn{1}{l|}{-} & - & \cref{subsec:lower} \\ \hline
\end{tabular}
\end{adjustbox}
\end{center}
\end{table*}

\section{Discussion}\label{sec:discussion}
In \cref{sec:jax}, we propose Masked DP-SGD which allows training with JAX with Poisson subsampling. However, we do not use any optimized clipping operation like GhostClipping or others (see \cref{tab:algo}) that yield higher throughput for certain types of layers and have been implemented with PyTorch. We see it as an interesting future direction to combine our Masked DP-SGD that enables training in JAX without recompiling with more efficient clipping implementations.

In \cref{fig:ThrvsBS_JAX_PyTorch} we saw that out JAX implementation had a
smaller gap in throughput between the small and large physical batches compared to for example the Opacus. We suspect this is due to some internal optimization that JAX is able to perform during the compilation for the accumulation step. It remains an interesting challenge for future research to see if further optimization of the throughput would be possible in JAX by more efficient use of resources.\looseness-1

Our JAX implementation also did not benefit from moving from FP32 to TF32, unlike Opacus. This is likely due to implementation-specific JAX choices, e.g.,~we upscale each of the feature vectors inside the loss function to fit the upscaled images to the VRAM. As we repeat this operation for each sample, we effectively waste some GPU cycles. As the FP32 yields significantly lower throughput, this bottleneck is not as visible in the FP32 comparison between JAX and Opacus. However, when moving to the TF32, the per-example gradient computation becomes faster, and the bottlenecks such as the image upscaling start to play a larger role in the throughput. This explains how in our NLP experiments, which avoid such data transformations with pre-tokenized inputs, outperforms Opacus at both precisions. Addressing the data loading could help with these limitations as a direction for future work.

In our comparison of the two use cases, computer vision and NLP, for DP training, we observed that the NLP case benefits significantly more from the use of TF32 with lower precision. The advantage can be attributed to the data dimensionalities and the use of a fixed max length, where all the vectors maintain the same shape.

In \cref{sec:multgpu}, we observe that DP PyTorch (using opacus) scales better than non-DP training.
We expect this finding to be generalizable to other frameworks, as the underlying reason for the different scaling behavior is that the compute is the dominating bottleneck for longer in DP training than without DP, where the network becomes a bottleneck already when communicating among four nodes with four GPUs each. However, training with many nodes requires sophisticated tooling that manages the communication and data loading. PyTorch supports this naively with \verb|DistributedDataParallel|, which utilizes the PyTorch DP library opacus, which has implemented a custom version for DP training. Throughout our experiments, we observed that implementing distributed training with JAX is more challenging, particularly when communicating between multiple nodes. However, we show that in the one-node setting, our method scales as expected, almost linearly (see \cref{fig:GPUs_wtf32_v2}). To scale like PyTorch across multiple nodes, JAX would also require a native implementation of data loading, which it currently does not support.\looseness-1

\section{Conclusion}

We summarize the lessons learnt in \cref{tab:lessons_learnt}. While DP-SGD is significantly more costly than non-private training, we identified feasible speed-ups that are often easy to apply but have some drawbacks. These are: 
    \begin{itemize}[leftmargin=*]  
    \item More efficient implementations of DP-SGD which additionally decrease the memory footprint (allowing for training larger models). However, these implementations are not as mature as Opacus and do not support all neural network layers (yet).
    \item JAX lacks a comprehensive DP-SGD implementation like PyTorch and exhibits greater variability in execution times. Although JAX processes samples faster than PyTorch, it loses the advantage through frequent re-compilations when utilizing proper Poisson sampling. We present an efficient DP-SGD implementation with JAX called Masked DP-SGD. It leverages JAX advantages in compilation and efficient sample processing, while adhering to Poisson subsampling requirements for correct privacy accounting. By avoiding frequent recompilation, we mitigate execution time variability and enhances efficient performance. It is future work to benefit from both optimizations, our proposed method and efficient clipping.
    \item Lower Precision using TF32 which increases throughput, but the implications on the theoretical guarantees of DP-SGD need to be explored in future work.
\end{itemize}
Finally, we found that distributed computing using DP-SGD scales better than non-private training and allows for fast training of models.

\section*{Acknowledgments}

This work was supported by the Finnish Ministry of Education and Culture and CSC - IT Centre for Science (Decision diary number OKM/10/524/2022), the Research Council of Finland (Flagship programme: Finnish Center for Artificial Intelligence, FCAI, Grant 356499 and Grant 359111), the Strategic Research Council at the Research Council of Finland (Grant 358247) as well as the European Union (Project 101070617). Sebastian Rodriguez Beltran was additionally supported by the EuroHPC Joint Undertaking and its members including top-up funding by the Finnish Ministry of Education and Culture. Niki Loppi contributed under the NVIDIA AI Technology Center (NVAITC) Finland program. Views and opinions expressed are however those of the author(s) only and do not necessarily reflect those of the European Union or the European Commission. Neither the European Union nor the granting authority can be held responsible for them. The authors wish to thank the CSC – IT Center for Science, Finland for supporting this project with computational and data storage resources. The authors acknowledge the research environment provided by ELLIS Institute Finland. We thank Talal Alrawajfeh for helpful discussions regarding implementing DP-SGD with JAX.

\section*{LLM usage considerations}

\begin{itemize}
    \item \textbf{Originality} For this work, we used LLM's for correcting grammar and clarifying some parts of the text. It was not used for the entire text of the paper, and the entire content is original from us.
    \item \textbf{Transparency} No idea was generated by an LLM.
    \item \textbf{Responsibility} Our work aims to provide an efficient implementation of a hard computational problem in training deep learning models under Differential Privacy. We hope to contribute to the democratization of models under privacy, as they become prohibitively expensive in compute and footprint.
\end{itemize}

\bibliographystyle{plainnat}
\bibliography{references}
\appendix

\setcounter{figure}{0} 
\renewcommand\thefigure{A.\arabic{figure}}
\setcounter{table}{0}
\renewcommand{\thetable}{A\arabic{table}}
\setcounter{equation}{0}
\renewcommand{\theequation}{A\arabic{equation}}

\section{Hyperparameters}
\label{sec:app-hyperparameters}

We use the computer vision hyperparameters obtained on request from \citet{Tobaben2023Efficacy}. The hyperparameters for all models are in \cref{tab:hypers}. Even though model utility is not the main objective in this work, in the non-private case, the learning rate is suboptimal. By changing it to 0.00027 we see an accuracy improvement, therefore the one we are using. We fine-tune all parameters at $\epsilon=8$ and $\delta=2.04e^{-5}$. The noise multiplier is adjusted to the number of steps to reach the DP budget at the end of training. For the NLP case in \cref{subsec:nlp}, we freeze the embedding layer and train the rest of the network. The learning rate comes from \citep{NLUBERT}.

\begin{table}[!h]
    \caption{Hyperparameters used for each model architecture.}
    \label{tab:hypers}
    \begin{center}
        \begin{small}
            \begin{sc}
                \begin{tabular}{lcc}
                    \toprule
                    Model & Learning Rate & Max Grad Norm \\
                    \midrule
                    ViT & 0.00031 & 4.63\\
                     ResNet & 0.00098 & 6.53\\
                     Bert & 0.00002 & 1 \\
                    \bottomrule
                    \end{tabular}
            \end{sc}
        \end{small}
    \end{center}  
\end{table}

\section{Additional Results}

This section provides additional figures that supplement the findings in the main text.

In \cref{tab:accuracy_comparison} we evaluate the effects of lower precision on accuracy. We evaluate over three different seeds, and different datasets. The throughput is consistent across datasets and by using lower precision, it is possible to double the samples per second. The differences between the two precisions have a mean of $\approx -4.3 \times 10^{-5}$ and a standard deviation of $\approx 4.2 \times 10^{-4}$. We cannot reject the null hypothesis that the accuracies have the same mean (p-value $\approx 0.74$). The use of a lower precision mode has a negligible effect on the accuracy.
\begin{table*}[ht]
    \centering
    \caption{Accuracy and throughput comparison of ViT Base Model for private training with Opacus, on FP32 and TF32 precision modes across two datasets with varying shot counts. We show the number of epochs (E), shots or number of examples per class (S) and the learning rate.}
    \begin{center}
        \begin{small}
            \begin{sc}
            \begin{tabular}{l|c|c|c|c|c|c|c|c|c}
        \toprule
        Dataset & E & S & lr & Throughput FP32 & Throughput TF32 & Acc FP32 & Acc TF32 & Std FP32 & Std TF32 \\
        \midrule
        CIFAR-10 & 18 & 250 & 0.000710 & 57.317226 & 117.987103 & 0.919950 & 0.923300 & 0.008273 & 0.008485 \\
        & 6 & 100 & 0.000758 &  56.950525 & 114.012746 & 0.956275 & 0.956367 & 0.001504 & 0.001850 \\
        \hline
        SVHN & 23 & 250 & 0.00098 & 57.279881 & 117.286507 & 0.610933 & 0.611321 & 0.009724 & 0.009848 \\
        & 23 & 500 & 0.00098 & 57.352352 & 118.368599 & 0.806301 & 0.806438 & 0.006607 & 0.007316 \\
        \hline
    \end{tabular}
            \end{sc}
        \end{small}
    \end{center}
    
    \label{tab:accuracy_comparison}
\end{table*}

\begin{table}[!h]
    \caption{Mean accuracy for IMDB test set for the BERT models on A100 after training for three epochs. The use of TF32 as a lower precision mode does not affect the model's utility. We use $\epsilon = 8$.}
    \label{tab:acc_table_bert}
    \begin{center}
        \begin{small}
            \begin{sc}
                \begin{tabular}{lcc}
                    \toprule
                    Clipping Mode & Test Accuracy \\
                    \midrule
                    Opacus & 0.70752 \\
                    Opacus/TF32 & 0.70516 \\
                    Masked DP-SGD & 0.7598\\
                    Masked DP-SGD / TF32 & 0.7600 \\
                    \bottomrule
                    \end{tabular}
            \end{sc}
        \end{small}
    \end{center}  
\end{table}

\begin{table}[!h]
    \caption{Mean accuracy for CIFAR-100 test set for each clipping mode for the ViT models on A100 after training for 92 epochs. All use the ViT hyperparameters from \cref{tab:hypers}. We use $\epsilon = 8$. }
    \label{tab:acc_table_full}
    \begin{center}
        \begin{small}
            \begin{sc}
                \begin{tabular}{lcc}
                    \toprule
                    Clipping Mode & Test Accuracy \\
                    \midrule
                    Opacus &  0.8764\\
                    Opacus/TF32 &   0.8762 \\
                    Masked DP-SGD &  0.9013 \\
                    PV-ghost & 0.8781 \\
                    BK-ghost & 0.8755 \\
                    \bottomrule
                    \end{tabular}
            \end{sc}
        \end{small}
    \end{center}  
\end{table}

\begin{figure}

\begin{center}
    \includegraphics[width=\columnwidth]{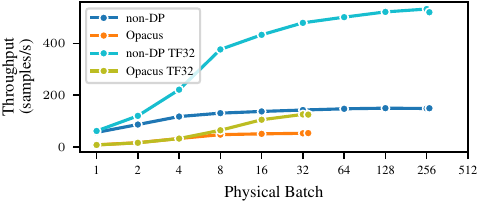}
    \hfill
     \includegraphics[width=\columnwidth]{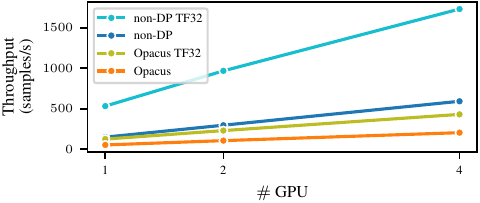}
   \caption{Combining distributed training with the use of lower precision TF32 for the ViT base model on A100.}
   \label{fig:GPUs_wtf32}%
\end{center}

\end{figure}

In \cref{fig:GPUs_wtf32} we show the progression of throughput in relation to the physical batch size, for Opacus and non-DP training, for both precision modes. We see that they present the same behavior as in \cref{fig:ThrvsBS_JAX_PyTorch}. TF32 also plateaus at an optimum without further gains.

\begin{figure}
    \begin{center}
    \centerline{\includegraphics[width=\columnwidth]{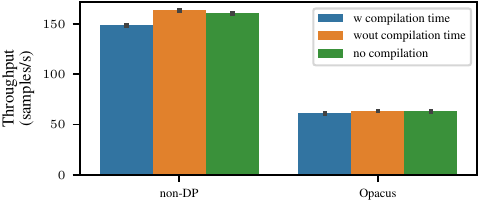}}
         \caption{Torch compilation experiments on A100 using ViT Base with maximum physical batch size for each mode. PyTorch 2 offers model compilation for potential speed-ups, but our tests showed no improvements compared to JAX compilation. PyTorch 2 attempts to compile but defaults to NVIDIA kernels and optimizations, resulting in unchanged throughput. Including initial compilation time leads to worse performance due to the time PyTorch spends trying to compile before defaulting. Disregarding the time where PyTorch tries to compile (wout compilation time), leads to nearly the same throughput to the non-compiling version.}
        \label{fig:compilation_torch}
    \end{center}
\end{figure}

\begin{figure}
    \begin{center}
        \centerline{\includegraphics[width=\columnwidth]{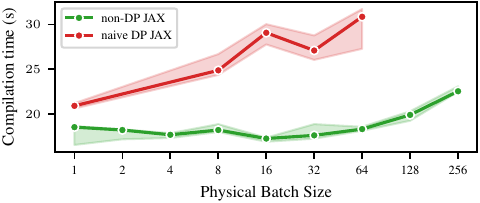}}
        \caption{Compilation time in seconds as a function of the physical batch size for JAX naive experiments for the ViT Base model on A100. The estimator is the median and the error bars are the $95\%$ confidence interval using bootstrapping.}
        \label{fig:compilation_jax}
    \end{center}
\end{figure}

\begin{figure}
    \begin{center}
        \centerline{\includegraphics[width=\columnwidth]{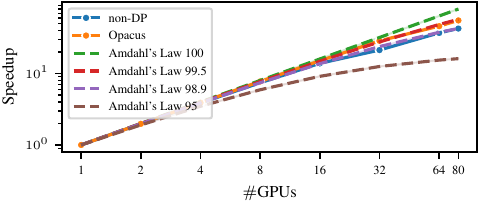}}
        \caption{Comparison between the throughput in our experiments and the theoretical Amdahl's Law. Both axis are in log scale. In the distributed setting, private training achieves a 99.5 $\%$ of parallel processing, with a 50 times speed up than single processing. }
        \label{fig:amdahls_law_puhti_log}
    \end{center}
\end{figure}

\begin{figure}
    \centering
    \includegraphics[width=\columnwidth]{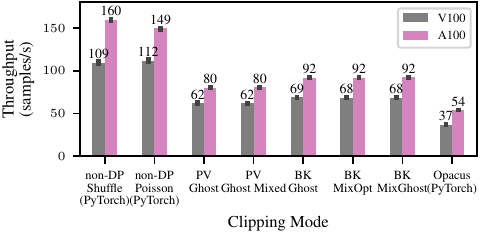}
    \caption{Throughput using the maximum batch size for each clipping algorithm and each GPU, for the ViT Base model, including two sampling methods for the non-DP baseline. Only shuffling the dataset at each epoch, or with Poisson subsampling.}
    \label{fig:shuffle}
\end{figure}

In \cref{fig:shuffle} we added another sampling method, to the previous \cref{fig:thrs_gpus_comp}. In it, we try a non-DP baseline shuffling the dataset instead of Poisson subsampling. We still use logical/physical batch size partitions. Our private Opacus baselines have a relative slowdown of $\times2.78$ compared to non-DP with Poisson subsampling. The relative slowdown compared to shuffling non-DP is $\times 3$ for the A100 GPU. Throughout the experiments and results we compare to the non-DP Poisson subsampling baseline.

\clearpage
\section{Poisson subsampling proof}\label{sec:poisson_proof}

We provide the following proof for \cref{lemma:poisson}:

\begin{proof} \label{proof:ofl1}
    With subsampling rate $q$, Poisson subsampling goes through each element of the 
    data set $D$ and with probability $q$ includes the element in the batch. Hence the
    batch size will be a $\mathrm{Binom}(N, q)$ distributed random variable, and each 
    batch of size $b$ will have a uniform probability of being sampled from the set of all the batches
    of size $b$. Therefore for any batch size $m$ sampling a particular set $B_{m, k}$ has a probability 
    \begin{align}
        \Pr(\mathrm{Poisson}(D, q) = B_{m, k}) = q^m \left( 1 - q  \right)^{N - m}. 
    \end{align}
    Now, it is easy to see, that if we first sample the batch size $b$ from $\mathrm{Binom}(N, q)$
    and finally we release $B_{b, k} \sim \mathrm{WOR}(D, b)$ we have
    \begin{align}
        \Pr(B = B_{m, k}) &= \Pr(B = B_{m, k} \mid b=m) \Pr(b = m) \\
                          &= \Pr(\mathrm{WOR}(D, m) = B_{m,k}) \Pr(b = m) \\
                          &= \binom{N}{m}^{-1} \binom{N}{m} q^m (1-q)^{N-m} \\
                          &= q^m (1-q)^{N-m}.
    \end{align}
    Hence we see that the probability of sampling any particular batch is the same between the two sampling procedures, and hence the claim holds.
\end{proof}

We outline the following detailed proof for the WOR sampling:
    
    \begin{itemize}
        \item Denote the upscaled batch size with $b_+$, the actual batch size with $b$ and the difference of the two with $\Delta(b) = b_+ - b$. 
        \item Sampling a random batch $B_+ \sim WOR(D, b_+)$ can be split into the following steps:
        
        \begin{enumerate}
            \item Sample $B \sim WOR(D, b)$
            \item Sample $B_{\Delta} \sim WOR(D \setminus B, \Delta(b))$
            \item Return $B_+ = B_{\Delta} \cup B$.
        \end{enumerate}
    
        \item In our masked DP-SGD, gradients for the $B_{\Delta}$ get masked away before gradient accumulation. 
        \item From the above procedure, it is easy to see that the samples that contribute to the accumulation, are sampled from $WOR(D, b)$. 
        \item As shown in our \cref{lemma:poisson}, the masked sampling procedure together with the sampling of the batch size from the Binomial distribution corresponds to the Poisson subsampling.
    \end{itemize}
    
    \textbf{Why the sampling can be split into three steps?} The WOR sampling a batch of $k$ samples from a data set $D$ can be iteratively done by repeating following steps until you have obtained $k$ samples:
    \begin{enumerate}
        \item Sample $x$ uniformly from $D_t$ and include $x$ in the batch
        \item Set $D_{t+1} = D_t \setminus {x}$, with $D_0 = D$.
    \end{enumerate}
    
    Therefore, it is easy to see that for integers $k_1, k_2 \leq k$ s.t. $k_1+k_2 = k$, sampling from $WOR(D, k)$ is equivalent to first sampling $B_1 \sim WOR(D, k_1)$ and $B_2 \sim WOR(D \setminus B_1, k_2)$ and finally releasing $B = B_1 \cup B_2$.

\end{document}